\documentclass[conference]{IEEEtran}
\IEEEoverridecommandlockouts
\usepackage{cite}
\usepackage{listings}
\usepackage{amssymb}
\usepackage{bm}
\usepackage{amsmath,amssymb,amsfonts}
\usepackage{algorithmic}
\usepackage{graphicx}
\usepackage{enumerate}
\usepackage{textcomp}
\usepackage{xcolor}
\usepackage{multirow}
\usepackage{url}

\newcommand{\revmagenta}[1]{#1}
\newcommand{\revred}[1]{#1}
\newcommand{\revblue}[1]{#1}
\newcommand{\revcyan}[1]{#1}
\newcommand{\revorange}[1]{#1}
\newcommand{\revgreen}[1]{#1}

\def\BibTeX{{\rm B\kern-.05em{\sc i\kern-.025em b}\kern-.08em
    T\kern-.1667em\lower.7ex\hbox{E}\kern-.125emX}}

\newsavebox{\codeboxbox}
\newenvironment{codebox}{%
  \vspace{0.5ex}\noindent\hspace*{0.5cm}%
  \begin{lrbox}{\codeboxbox}%
  \begin{minipage}{\dimexpr\columnwidth-0.5cm-2\fboxsep-2\fboxrule\relax}%
}{%
  \end{minipage}%
  \end{lrbox}%
  \fbox{\usebox{\codeboxbox}}%
  \par\vspace{0.5ex}%
}

\title{\revmagenta{Improving HPC Code Generation Capability of LLMs via Online Reinforcement Learning with Real-Machine Benchmark Rewards}}

\author{
\noindent
\parbox[t]{0.3\textwidth}{\centering
Ryo Mikasa\\
Department of Computer Science,\\
Nagoya University
}
\hfill
\parbox[t]{0.3\textwidth}{\centering
Shun-ichiro Hayashi\\
Graduate School of Informatics,\\
Nagoya University
}
\hfill
\parbox[t]{0.3\textwidth}{\centering
Daichi Mukunoki\\
Information Technology Center,\\
Nagoya University
}
\\[0.6em]
\hfill
\parbox[t]{0.45\textwidth}{\centering
Tetsuya Hoshino\\
Information Technology Center,\\
Nagoya University
}
\hfill
\parbox[t]{0.45\textwidth}{\centering
Takahiro Katagiri\\
Information Technology Center,\\
Nagoya University
}
\hfill
}

\begin{document}
\maketitle

\begin{abstract}
\revmagenta{Large language models (LLMs) have demonstrated strong code generation capabilities, yet the runtime performance of generated code is not guaranteed, and there have been few attempts to train LLMs using runtime performance as a reward in the HPC domain. We propose an online reinforcement learning approach that executes LLM-generated code on a supercomputer and directly feeds back the measured runtime performance (GFLOPS) as a reward. We further introduce a Staged Quality-Diversity (SQD) algorithm that progressively varies the permitted optimization techniques on a per-problem basis, enabling the model to learn code optimization from diverse perspectives. We build a distributed system connecting a GPU training cluster with a CPU benchmarking cluster, and train Qwen2.5 Coder 14B on a double-precision matrix multiplication task using Group Relative Policy Optimization (GRPO). Through two experiments, we show that reinforcement learning combining runtime performance feedback with staged optimization can improve the HPC code generation capability of LLMs.}
\end{abstract}


\section{Introduction}
High-Performance Computing (HPC) is essential for scientific computing and large-scale data analysis. While performance optimization techniques such as thread-parallelization with OpenMP, Single Instruction Multiple Data (SIMD) instructions, and cache/register blocking can approach theoretical peak performance, they require deep knowledge of the target architecture and memory hierarchies. Implementing such optimizations typically demands substantial expert effort.
\par 

Large language models (LLMs) have recently shown strong code generation capabilities~\cite{codellama}. They are deep neural networks trained on large-scale text and code data, and can generate program code from natural-language instructions. However, most LLMs are trained to maximize functional correctness (e.g., passing tests) rather than runtime performance. \revmagenta{While supervised fine-tuning can adapt models to specific domains, it requires high-quality reference data, which is scarce for HPC optimization. Reinforcement learning (RL) offers an alternative by learning directly from performance feedback without reference solutions.} In HPC, two functionally equivalent implementations can differ by orders of magnitude in speed, making performance-aware code generation a crucial challenge.
\revred{Recent HPC-oriented studies highlight both the promise and the limitations of LLMs for performance-critical code: HPC-Coder~\cite{hpccoder} and ParEval~\cite{pareval} report meaningful gains on parallel code generation, while benchmarks such as Mercury~\cite{mercury} and KernelBench~\cite{kernelbench} focus on measuring efficiency rather than training for it.}
\revred{At the same time, systems like ChatHPC~\cite{chathpc} demonstrate end-to-end HPC assistants with model training in the loop, suggesting that performance-aware learning pipelines are a natural next step for HPC code generation.}

We address this by directly evaluating LLM-generated code on a supercomputer and using measured GFLOPS as the reward signal for Group Relative Policy Optimization (GRPO)~\cite{deepseekmath}. This avoids the need for hand-crafted reference solutions and enables learning from performance feedback alone.

To demonstrate this concept, we use \revmagenta{matrix multiplication ($C = A \times B$)}, one of the most fundamental kernel operations in HPC, as an example. We use Qwen2.5 Coder 14B~\cite{qwen25coder} (details in Section~\ref{sec:background}) as the base model for \revblue{reinforcement learning (RL)}. We conduct two experiments with different system setups. Experiment 1 uses a single fixed prompt and explores hyperparameters. Experiment 2 builds a more advanced system with staged optimization and a tree-based data evolution mechanism. Experiment 1 evaluates 12 configurations across compiler optimization level, learning rate, and KL (Kullback--Leibler) penalty. Experiment 2 proposes the Staged Quality-Diversity (SQD) algorithm and analyzes learning dynamics across six configurations.

Our contributions are as follows.
\begin{enumerate}
    \item A new RL framework that improves HPC code generation using execution-based benchmarks
    \item A systematic analysis of how compiler optimization and learning rate affect learning dynamics
    \item The SQD algorithm, which enables the model to autonomously explore diverse optimization paths and learn error recovery without human intervention, aiming to deepen its understanding of code semantics.
\end{enumerate}

\revmagenta{The rest of this paper is organized as follows. Section~\ref{sec:background} reviews background on LLM training and GRPO. Section~\ref{sec:proposed} presents the proposed method, including the reward design, SQD algorithm, and distributed training architecture. Section~\ref{sec:experiment} reports experimental results, and Section~\ref{sec:related} discusses related work. Section~\ref{sec:conclusion} concludes.}

\section{Background}
\label{sec:background}

\subsection{Training Large Language Models}

LLMs are deep neural networks based on the Transformer architecture~\cite{transformer}, pre-trained to predict the next token from large-scale text and code corpora. Here, a token corresponds to a word or symbol unit. Pretraining lets LLMs acquire programming language syntax as well as common coding patterns, while task-specific performance is often suboptimal.

\revmagenta{In this work, we use Qwen2.5 Coder 14B~\cite{qwen25coder} as the base model. Qwen2.5 Coder is a code-specialized LLM developed by Alibaba and continually pre-trained on 5.5 trillion tokens of code corpora. It supports 92 programming languages and can process long contexts of up to 131,072 tokens. However, it has not been specifically trained for HPC optimization; we therefore apply GRPO-based RL to improve its matrix multiplication optimization capability.}

Two common adaptation methods are supervised fine-tuning and \revblue{reinforcement learning (RL)}. \revmagenta{Supervised fine-tuning requires high-quality reference solutions as training data, which are scarce for HPC optimization. RL, in contrast, does not require such reference data; it updates the model parameters to increase the likelihood of outputs that receive higher rewards. In this work, we use runtime performance (GFLOPS) as the reward, making RL a natural fit.}

\subsection{RL for LLMs: From PPO to GRPO}

\revmagenta{Proximal Policy Optimization (PPO)~\cite{ppo} is the standard RL algorithm for LLM training. PPO constrains policy updates so that the new policy does not diverge too far from the old one, thereby stabilizing learning. However, PPO requires a critic model that approximates the value function, which is typically comparable in size to the LLM itself; this significantly increases memory usage and computational cost.}

\revmagenta{Direct Preference Optimization (DPO)~\cite{dpo} eliminates both the reward model and the critic by directly optimizing the policy from preference data. While computationally efficient, DPO is an offline method that relies on pre-collected preference datasets and does not interact with the environment during training. In our setting, where benchmark results are obtained in real time and fed back as rewards, an online RL method is required.}

\revmagenta{Group Relative Policy Optimization (GRPO)~\cite{deepseekmath} removes the critic and instead compares multiple samples generated for the same prompt. For each prompt, $G$ outputs are sampled and treated as a group; the group mean reward serves as the baseline, removing the need for a separate critic model. This substantially reduces memory usage compared with PPO, allowing more resources to be devoted to the model itself. GRPO was adopted in DeepSeek-R1~\cite{deepseekr1} and has shown strong results in mathematical reasoning tasks. Because GRPO evaluates samples relatively within a group, the model learns to produce ``better code than its other attempts,'' which aligns naturally with our objective of maximizing runtime performance.}

For prompt $x$, the model $\pi_\theta$ samples $G$ outputs $\{y_1, y_2, \ldots, y_G\}$, which are treated as a group. The advantage of each sample $y_i$ is computed from its relative reward within the group:
\begin{equation}
A_i = \frac{R_i - \mu_R}{\sigma_R + \epsilon}
\end{equation}
where $R_i$ is the reward of sample $y_i$, $\mu_R$ and $\sigma_R$ are the group mean and standard deviation of rewards, and $\epsilon$ is a small constant for numerical stability. \revmagenta{A positive advantage indicates that the sample performed better than the group average.}
The GRPO objective maximizes expected reward while controlling divergence from the original model:
\begin{align}
\mathcal{L}_{\text{GRPO}}(\theta) = \mathbb{E}_{x,y} \bigl[ &\min\left( \rho_t A, \text{clip}(\rho_t, 1{-}\delta, 1{+}\delta) A \right) \notag \\
&- \beta \cdot D_{\text{KL}}(\pi_\theta \Vert \pi_{\text{ref}}) \bigr]
\end{align}
\revmagenta{where $A$ is the advantage computed above, $\rho_t = \pi_\theta(y|x) / \pi_{\text{old}}(y|x)$ is the importance ratio between the current policy $\pi_\theta$ and the policy $\pi_{\text{old}}$ used to generate the samples, $\text{clip}(\cdot, 1{-}\delta, 1{+}\delta)$ is the clipping function (as in PPO) that prevents excessively large policy updates, and $\delta$ is the clipping range (e.g., 0.2). The KL divergence term $D_{\text{KL}}(\pi_\theta \Vert \pi_{\text{ref}})$ measures the distributional distance between the updated model $\pi_\theta$ and the reference model $\pi_{\text{ref}}$ (i.e., the original pre-trained model), and $\beta$ is the KL penalty coefficient that controls the strength of this regularization. A larger $\beta$ enforces more conservative updates, while $\beta=0$ leaves exploration unconstrained.}

\section{Proposed Method}
\label{sec:proposed}

We train an LLM to generate high-performance HPC code using real benchmark performance as the reward. This differs from correctness-only reward signals and enables optimization over continuous performance metrics. \revmagenta{This section describes the reward design, staged constraints, SQD algorithm, distributed training architecture, and online training loop.}

\subsection{Reward Function Design}

\revmagenta{This study targets double-precision matrix multiplication $C = A \times B$, where $A$, $B$, and $C$ are matrices of dimensions $M \times K$, $K \times N$, and $M \times N$, respectively. We set $M = N = K = 256$.} We use a multi-level reward function:
\begin{equation}
 r = \begin{cases}
-100 & \text{(format/library violation)} \\
-40 & \text{(compile error)} \\
-30 & \text{(timeout)} \\
-20 & \text{(runtime error)} \\
-10 & \text{(verification error)} \\
\text{GFLOPS} & \text{success}
\end{cases}
\label{eq:reward}
\end{equation}

\revmagenta{Each penalty category corresponds to a stage in the code execution pipeline:
\begin{itemize}
    \item \textbf{Format violation} ($-100$): The generated output cannot be parsed as a valid C function with the expected signature, preventing extraction of compilable code.
    \item \textbf{Library violation} ($-100$): The code references prohibited external libraries or functions (e.g., calling BLAS routines directly), which would bypass the optimization task.
    \item \textbf{Compile error} ($-40$): The extracted code fails to compile due to syntax errors or undeclared identifiers.
    \item \textbf{Timeout} ($-30$): The code compiles but does not terminate within the time limit of 8 seconds, typically due to infinite loops or excessive computation.
    \item \textbf{Runtime error} ($-20$): The code compiles and starts execution but crashes (e.g., segmentation fault).
    \item \textbf{Verification error} ($-10$): The code runs to completion but produces numerically incorrect results, as described below.
\end{itemize}}

\revcyan{The hierarchy of penalties reflects the code generation pipeline: a verification error ($-10$) is preferred over a runtime error ($-20$) because it indicates that the code is memory-safe and structurally valid enough to complete execution, even if the logic is flawed. These penalty values were determined empirically.}

Verification compares the output matrix with OpenBLAS\footnote{\url{https://www.openblas.net/}} (cblas\_dgemm) at 105 sampled points: the four corners, center, and 100 random positions. If any relative error exceeds $10^{-6}$, the code is marked incorrect. Only codes passing this verification are used to report performance results, while incorrect or failed runs receive penalties during training. \revmagenta{The staged penalties encourage progression from syntactically correct code to fully correct and performant implementations: a program that runs but produces incorrect output receives a milder penalty than one that fails to compile, letting the model learn how far it progressed in the execution pipeline.}

The success reward differs across experiments. Experiment 1 uses raw GFLOPS to maximize absolute performance. Experiment 2 uses a group-normalized reward ($\text{GFLOPS}_{\text{generated}} / \text{GFLOPS}_{\text{max}}^{(\text{group})} \times 100$). \revmagenta{In stages where fewer techniques are permitted, the achievable absolute performance is lower, so raw values would weaken the performance-difference signal in GRPO's relative comparison. This normalization ensures that performance differences within each stage are appropriately reflected.}

\subsection{Staged Optimization Constraints}

Experiment 2 defines six optimization stages (Table~\ref{tab:stages}) and forbids certain techniques at each stage using regex detection. The five techniques are as follows.
\begin{enumerate}
    \item Register blocking
    \item Cache blocking
    \item SIMD instructions (AVX-512)
    \item OpenMP thread-parallelization
    \item Memory prefetching
\end{enumerate}

\begin{table}[t]
\caption{Optimization stage definitions}
\label{tab:stages}
\centering
\begin{tabular}{c|l|l}\hline
Stage & Allowed techniques & Violations detected \\ \hline
1 & Register blocking & SIMD, OMP, Pref \\
2 & +Cache blocking & \revmagenta{SIMD, OMP, Pref} \\
3 & +SIMD (AVX-512) & \revmagenta{OMP, Pref} \\
4 & +OpenMP & \revmagenta{Pref} \\
5 & +Prefetch & None \\
6 & All techniques & None \\ \hline
\end{tabular}
\end{table}

\subsection{SQD Algorithm}

We propose the Staged Quality-Diversity (SQD) algorithm for Experiment 2, formulating the optimization as a search tree (Fig.~\ref{fig:sqd}). 
\revorange{Conceptually, SQD is designed to enable the LLM to autonomously generate diverse optimization strategies that serve as seeds for further exploration. The model is exposed to a wide spectrum of code states, from initial broken implementations to highly optimized kernels, so that it can learn both optimization logic and error recovery mechanisms without relying on manually curated training cases. This approach is intended to foster a deeper understanding of code semantics by traversing the solution space more broadly than standard fine-tuning methods.}

To achieve this, SQD narrows the search space in a controlled way using stage-specific constraints. In each iteration, nodes are selected for expansion as follows.
\begin{enumerate}
    \item \revmagenta{Deepening: the top 40\% of nodes by performance within a stage are advanced to the next stage}
    \item \revmagenta{Improving: the middle 30\% generate new variants at the same stage}
    \item \revmagenta{Repairing: the remaining 30\% that produced errors generate corrected versions}
\end{enumerate}
These ratios were determined empirically to balance exploration across stages.
Selected codes are expanded by the LLM, benchmarked, and immediately used for GRPO updates. This generate--evaluate--learn cycle enables online RL that effectively couples exploration with learning.

\begin{figure}[t]
\begin{codebox}
\begin{lstlisting}
Input: seed code x_seed, iterations T
Pool(0) <- {stage1: {x_seed}}  // stage -> codes
for t = 1..T:
  P <- {x_seed}
  for each stage s in Pool(t-1):
    P <- P + SelectTop(Pool(t-1)[s], 40%)
    P <- P + SelectMid(Pool(t-1)[s], 30%)
    P <- P + SelectErr(Pool(t-1)[s], 30%)
  Pool(t) <- {}  // reset each iteration
  for each node x in P:
    children <- pi.generate(x, G)
    Benchmark(children)
    pi <- pi.update(children)  // GRPO
    for each c in children:
      Pool(t)[c.stage] += {c}
\end{lstlisting}
\end{codebox}
\caption{SQD with online RL.}
\label{fig:sqd}
\end{figure}

\revmagenta{\subsection{Distributed Training Architecture}}

\begin{figure*}[t]
\centering
\revmagenta{\includegraphics[width=0.8\textwidth]{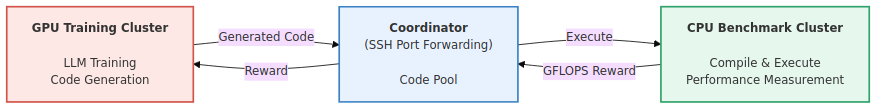}}
\caption{\revmagenta{Distributed training architecture: GPU training cluster and CPU benchmark cluster connected via a coordinator}}
\label{fig:system_architecture}
\end{figure*}

\revmagenta{Our method separates GPU-based training from CPU-based benchmarking (Fig.~\ref{fig:system_architecture}). The GPU training cluster handles LLM training and code generation, while the CPU benchmark cluster compiles and executes the generated code to measure runtime performance. A coordinator node manages data transfer and execution control between the two clusters via SSH port forwarding. This architecture decouples the training and evaluation environments, allowing flexible pairing of heterogeneous resources regardless of their physical location.}

\revmagenta{\subsection{Online GRPO Training Loop}}

\revmagenta{Each GRPO training step proceeds as follows.}
\begin{enumerate}
    \item \revmagenta{Load optimization prompts for the target problem}
    \item \revmagenta{Generate $G$ code samples per prompt on the GPU cluster ($G=64$ in our experiments)}
    \item \revmagenta{Transfer, compile, and benchmark each sample on the CPU cluster to compute rewards (Eq.~\ref{eq:reward})}
    \item \revmagenta{Compute group-relative advantages and update the model with GRPO}
\end{enumerate}
\revmagenta{Repeating this generate--evaluate--learn cycle enables online reinforcement learning from real performance feedback. The tight coupling between code generation and hardware evaluation is a key feature of our method, ensuring that the model continuously receives reward signals grounded in actual runtime measurements.}

\section{Experiments}
\label{sec:experiment}

\subsection{Experimental Setup}

Training used two nodes from Kyushu University supercomputer ``Genkai'' (node group B). Each node has Intel Xeon Platinum 8490H (60 cores $\times$ 2 sockets) and four NVIDIA H100 GPUs (eight total), with 1,024~GiB DDR5 memory and InfiniBand NDR 400~Gbps. The model was distributed with \revmagenta{tensor parallelism (TP=8)}.

Benchmarks ran on four nodes of Nagoya University supercomputer ``Flow'' Type II. Generated codes were dispatched across the four nodes in parallel for benchmarking. Each node has Intel Xeon Gold 6230 (20 cores $\times$ 2 sockets, 40 total), 384~GiB DDR4 memory, and AVX-512 support. The theoretical peak per socket is about 1.34~TFLOPS (20 cores $\times$ 2.1~GHz $\times$ 8 (AVX-512) $\times$ 2 (FMA: mul+add) $\times$ 2 (FMA units per core)), for 2.7~TFLOPS (2,688~GFLOPS) per node. Matrix size is $M=N=K=256$. We chose this size to keep each benchmark short because RL requires repeated measurements over many generated codes. \revmagenta{Even at this size, performance differences among generated codes span two orders of magnitude (e.g., 7 to 549 GFLOPS), providing a sufficiently rich reward signal for meaningful training.}
GCC 11.3.0 was used with \texttt{-march=native -fopenmp}, and optimization levels were -O0, -O1, or -O3 depending on the experiment.
\begin{itemize}
    \item \textbf{O0} (no optimization): Disables compiler optimization. This is used for debugging purposes, as the source code is converted almost as is into assembly. The quality of the source code is directly reflected in performance.
    \item \textbf{O1} (basic optimization): Performs optimizations that do not significantly increase compilation time, such as constant folding, dead code elimination, and basic register allocation optimizations. Does not perform automatic vectorization.
    \item \textbf{O3} (aggressive optimization): Applies all available optimizations, such as automatic vectorization (automatic generation of SIMD instructions), loop unrolling, function inlining, and loop structure transformation. High performance can sometimes be achieved even without the programmer explicitly specifying optimizations.
\end{itemize}
After one warm-up run, three runs were averaged to compute GFLOPS.

We use Qwen2.5 Coder 14B~\cite{qwen25coder} as the base model, Megatron-Core~\cite{megatron} as the distributed training framework, vLLM~\cite{vllm} as the inference engine for code generation, and ms-swift~\cite{swiftmlm} as the GRPO training wrapper.

\subsection{Experiment 1: Hyperparameter Search with Fixed Prompts}

\subsubsection{Overview}

Using eight fixed prompts (requesting all optimization techniques), we trained for about 200 steps across 12 configurations: compiler optimization level (O0/O3), learning rate \revmagenta{(LR;} 2e-7/5e-7), and KL penalty $\beta$ (0/0.001/0.01). Larger $\beta$ values constrain deviation from the pre-trained model.

\subsubsection{Results}

\begin{figure}[t]
\centering
\includegraphics[width=\columnwidth]{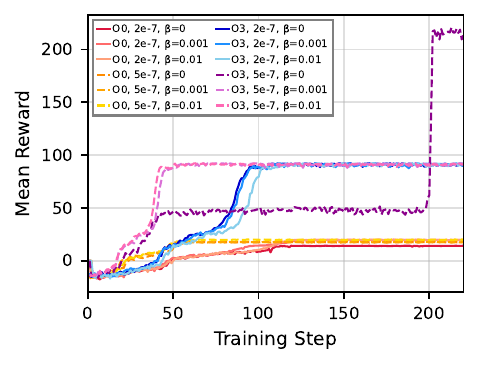}
\caption{Experiment 1: Mean reward across all settings. \revmagenta{Reward is raw GFLOPS for successful codes and a negative penalty for failures (Eq.~\ref{eq:reward}).} Solid lines are LR=2e-7; dashed lines are LR=5e-7.}
\label{fig:exp1_results}
\end{figure}

\begin{figure}[t]
\centering
\includegraphics[width=\columnwidth]{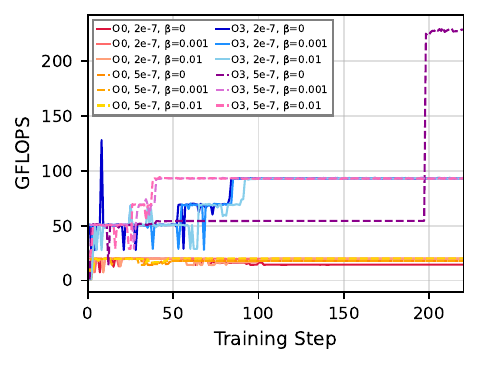}
\caption{Experiment 1: Max GFLOPS per step. Solid lines are LR=2e-7; dashed lines are LR=5e-7.}
\label{fig:exp1_max}
\end{figure}

Fig.~\ref{fig:exp1_results} shows mean reward, and Fig.~\ref{fig:exp1_max} shows max GFLOPS per step. O3 achieved final mean rewards of 70--78, while O0 remained at 7--15 (about 5--10$\times$ lower). The O3, LR=5e-7, $\beta=0$ configuration exhibited a jump from 54 to 225 GFLOPS around step 198. O3, LR=2e-7, $\beta=0$ reached 128 GFLOPS at step 8 but later regressed. With $\beta=0$ exploration continued, while $\beta \geq 0.001$ converged early.
The jump from 54 to 225 GFLOPS corresponds to roughly a 4x increase.
After reaching 128 GFLOPS, the O3, LR=2e-7, $\beta=0$ run fell back to around 50 GFLOPS and did not recover.
\revcyan{This suggests a complex interaction where model-learned optimizations (e.g., manual unrolling or intrinsics) may interfere with compiler heuristics. While -O0 provides a smooth reward landscape reflecting the model's raw capability, -O3 introduces a volatile landscape where minor code changes can break the compiler's auto-vectorization analysis, leading to abrupt performance drops.} The regression at lower learning rate implies that without sufficient update intensity, acquired optimization patterns may be forgotten.

\subsubsection{Discussion}

\revmagenta{Regarding compiler optimization, O3 reached a mean reward of 70--78 due to compiler auto-optimization, while O0 stayed at 7--15, highlighting the difficulty of correct SIMD usage. After the performance jump, patterns such as \texttt{collapse(2)} and i-k-j loop order emerged and stabilized. As for learning rate and KL penalty, LR=5e-7 retained high-performance patterns, while LR=2e-7 forgot them. With $\beta=0$, exploration continued and enabled discovery of high-performance patterns, whereas $\beta \geq 0.001$ converged early.}

\subsubsection{Generated Code Analysis}

We compare O3, $\beta=0$ code before and after training. Fig.~\ref{lst:exp1_before} shows a pre-training sample. It uses SIMD intrinsics but loads K-direction vectors from A and B, producing semantic errors and failing validation.

\begin{figure}[t]
\begin{codebox}
\begin{lstlisting}
#include <immintrin.h>
#include <omp.h>
#include <math.h>
#define BLOCK_SIZE 128
void matmul_optimized(double *A, double *B,
    double *C, int M, int N, int K) {
  #pragma omp parallel for
  for (int i = 0; i < M; i++)
    for (int j = 0; j < N; j++)
      C[i * N + j] = 0.0;
  #pragma omp parallel for collapse(3)
  for (int ib = 0; ib < M; ib += BLOCK_SIZE) {
    for (int jb = 0; jb < N; jb += BLOCK_SIZE) {
      for (int kb = 0; kb < K; kb += BLOCK_SIZE) {
        for (int i = ib; i < i_end; i++) {
          for (int j = jb; j < j_end; j++) {
            __m512d c0 = _mm512_loadu_pd(&C[i*N+j]);
            for (int k = kb; k < k_end; k += 8) {
              __m512d a = _mm512_loadu_pd(&A[i*K+k]);
              __m512d b = _mm512_loadu_pd(&B[k*N+j]);
              c0 = _mm512_fmadd_pd(a, b, c0);
            }
            _mm512_storeu_pd(&C[i*N+j], c0);
            }
          }
        }
      }
    }
  }
}
\end{lstlisting}
\end{codebox}
\caption{Experiment 1: Generated code before training (step 0, verification error).}
\label{lst:exp1_before}
\end{figure}

\revblue{Fig.~\ref{lst:exp1_after} shows the best post-training code, achieving 230.70 GFLOPS. The code does not employ explicit SIMD instructions; instead, it uses \texttt{collapse(2)} and scalar accumulators, and delegates vectorization to the compiler's auto-vectorization enabled by -O3, thereby achieving high performance. This result \revmagenta{suggests} that, through extensive code generation and parameter updates driven by execution feedback, the model empirically learned an optimization strategy in which relying on compiler auto-vectorization yields more stable and higher performance than complex manual SIMD optimizations.}

\begin{figure}[t]
\begin{codebox}
\begin{lstlisting}
#include <immintrin.h>
#include <omp.h>
#define BLOCK_SIZE 2048
void matmul_optimized(double *A, double *B,
    double *C, int M, int N, int K) {
  #pragma omp parallel for
  for (int i = 0; i < M; ++i)
    for (int j = 0; j < N; ++j)
      C[i * N + j] = 0.0;
  for (int mb = 0; mb < M; mb += BLOCK_SIZE) {
    int m_end = mb+BLOCK_SIZE>M ? M : mb+BLOCK_SIZE;
    for (int kb = 0; kb < K; kb += BLOCK_SIZE) {
      int k_end = kb+BLOCK_SIZE>K ? K : kb+BLOCK_SIZE;
      for (int nb = 0; nb < N; nb += BLOCK_SIZE) {
        int n_end = nb+BLOCK_SIZE>N ? N : 
                                        nb+BLOCK_SIZE;
        #pragma omp parallel for collapse(2)
        for (int i = mb; i < m_end; ++i) {
          for (int j = nb; j < n_end; ++j) {
            double cij = C[i * N + j];
            for (int k = kb; k < k_end; ++k)
              cij += A[i*K+k] * B[k*N+j];
            C[i * N + j] = cij;
          }
        }
      }
    }
  }
}
\end{lstlisting}
\end{codebox}
\caption{Experiment 1: Best post-training code (-O3, 230.70 GFLOPS).}
\label{lst:exp1_after}
\end{figure}

\revmagenta{\subsection{Experiment 2: SQD-Based Staged Optimization}}

\revmagenta{Experiment 1 showed that $\beta=0$ yields the best exploration and that runtime feedback can drive learning, but training with fixed prompts limits the diversity of generated code. To address this, Experiment 2 applies the SQD algorithm, which progressively varies optimization constraints to expose the model to a wider range of code states.}

\subsubsection{Overview}

With $\beta=0$ fixed, we ran SQD training up to iteration 9 across six configurations: compiler optimization level (O0/O1/O3) and learning rate (0/5e-7). LR=0 is a no-update baseline (SQD exploration only).

\subsubsection{Results}

\begin{figure}[t]
\centering
\includegraphics[width=\columnwidth]{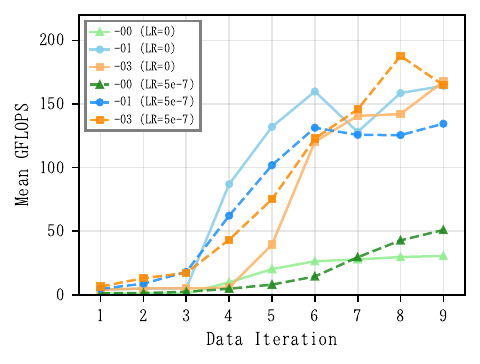}
\caption{Experiment 2: Mean GFLOPS across six settings. Solid lines are LR=0 (no learning); dashed lines are LR=5e-7 (with learning).}
\label{fig:exp2_mean}
\end{figure}

\begin{figure}[t]
\centering
\includegraphics[width=\columnwidth]{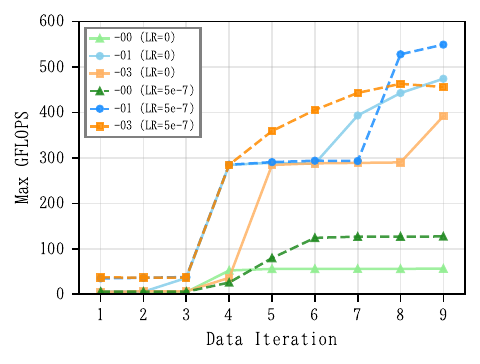}
\caption{Experiment 2: Max GFLOPS across six settings. Solid lines are LR=0; dashed lines are LR=5e-7.}
\label{fig:exp2_max}
\end{figure}

Fig.~\ref{fig:exp2_mean} shows mean GFLOPS, and Fig.~\ref{fig:exp2_max} shows max GFLOPS. Table~\ref{tab:exp2_summary} summarizes performance.
\revred{Compared with Experiment~1, staged SQD training yields clearer separation between LR=0 and LR=5e-7, indicating that learning (not just exploration) drives gains. The larger improvement at O0 than at O1 suggests that weaker compiler optimization leaves more headroom for RL to discover effective transformations.}

\begin{table}[t]
\caption{Performance comparison for Experiment 2 (up to iteration 9)}
\label{tab:exp2_summary}
\centering
\begin{tabular}{lrrr}\hline
Setting & Best GFLOPS & Peak ratio & Learning gain \\ \hline
O0, LR=0 & 57 & 2.1\% & -- \\
O0, LR=5e-7 & 127 & 4.7\% & +125\% \\
O1, LR=0 & 474 & 17.6\% & -- \\
O1, LR=5e-7 & 549 & 20.4\% & +16\% \\
O3, LR=0 & 392 & 14.6\% & -- \\
O3, LR=5e-7 & 463 & 17.2\% & +18\% \\ \hline
\end{tabular}
\end{table}

\subsubsection{Discussion}

\revmagenta{Regarding compiler optimization, O0 benefited most (+125\%, 57 to 127 GFLOPS). O1 achieved 549 GFLOPS (20.4\% peak, +16\%), and O3 improved by +18\% (392 to 463 GFLOPS). Strong compiler optimization can absorb minor code differences, limiting learning gains, while weaker optimization makes code quality more directly visible in performance. As for learning rate, with LR=0, O1 reached 474 GFLOPS and O3 reached 392 GFLOPS, showing strong prior knowledge in Qwen2.5 Coder. LR=5e-7 improved all settings without mode collapse. Compared with Experiment 1's maximum of 230.70 GFLOPS, Experiment 2 achieved 549 GFLOPS via SQD.}

\subsubsection{Generated Code Analysis}

Fig.~\ref{lst:o0_lr0} through Fig.~\ref{lst:o3_lr5e7} show the best code for each setting, and Table~\ref{tab:exp2_code_features} summarizes structural features. The matrix size is $M=N=K=256$, and the total 1.5~MB of three double-precision matrices fits in the 27.5~MB L3 cache.

\begin{table}[t]
\caption{Structural features of best-performing code per setting}
\label{tab:exp2_code_features}
\centering\small
\begin{tabular}{@{}l@{\hskip 6pt}r@{\hskip 6pt}r@{\hskip 6pt}r@{\hskip 6pt}r@{}}\hline
Setting & GFLOPS & Blocking & Threads & Prefetch \\ \hline
O0, LR=0 & 57 & 64 & collapse(2) & k+1 \\
O1, LR=0 & 474 & 64 & manual split & k+8 \\
O3, LR=0 & 392 & 64 & collapse(2) & none \\ \hline
O0, LR=5e-7 & 127 & 128 & manual split & k+1 \\
O1, LR=5e-7 & 549 & none & par. for & k+1 \\
O3, LR=5e-7 & 463 & 64 & manual split & block boundary \\ \hline
\end{tabular}
\end{table}

The best O1, LR=5e-7 code (549 GFLOPS) uses a simple triple loop without blocking, reflecting the fact that the problem fits in cache and blocking overhead hurts performance. O1, LR=0 (474 GFLOPS) uses 64\,\,x\,\,64 blocking and a read-modify-write C update; its overhead and extra memory traffic explain the performance gap. Many generated codes omit boundary checks, implicitly assuming matrix sizes are multiples of block sizes or SIMD widths (8). This indicates overfitting to the 256\,\,x\,\,256 training size\revgreen{. Although this limits immediate applicability, it successfully demonstrates that the model can learn complex optimization logic via RL. Generalizing} to arbitrary sizes \revgreen{through parameterized prompts or varied training inputs} remains future work.

\begin{itemize}\setlength{\leftmargin}{0pt}\setlength{\itemsep}{2pt}
\item \textbf{O0, LR=0 (57 GFLOPS):} Fig.~\ref{lst:o0_lr0}. 64\,x\,64 blocking, \texttt{collapse(2)} parallelization, 8 accumulators, prefetch (k+1).
\item \textbf{O1, LR=0 (474 GFLOPS):} Fig.~\ref{lst:o1_lr0}. Manual thread partitioning, 64\,x\,64 blocking, 8 accumulators, prefetch (k+8). Read-modify-write update of C.
\item \textbf{O3, LR=0 (392 GFLOPS):} Fig.~\ref{lst:o3_lr0}. 64\,x\,64 blocking, \texttt{collapse(2)} parallelization, 8\,x\,8 register blocking. Broadcasts A values and applies FMA against B vectors.
\item \textbf{O0, LR=5e-7 (127 GFLOPS):} Fig.~\ref{lst:o0_lr5e7}. 128\,x\,128 blocking, manual thread splitting, 8 accumulators, prefetch (k+1). With -O0, explicit SIMD is required.
\item \textbf{O1, LR=5e-7 (549 GFLOPS):} Fig.~\ref{lst:o1_lr5e7}. Simple triple loop with \texttt{parallel for} row partitioning, 8 accumulators, aligned loads, prefetch (k+1). Writes C directly without read-modify-write and achieves the best overall performance. The benchmark target is $C = A \times B$, so cumulative addition to C is unnecessary. O1, LR=0 employs a read-modify-write pattern originating from the BLAS GEMM pattern ($C = \alpha AB + \beta C$) in pre-training knowledge, introducing unnecessary memory accesses. With LR=5e-7, GRPO updates eliminated this redundant operation, demonstrating that RL can overcome pre-training bias.
\item \textbf{O3, LR=5e-7 (463 GFLOPS):} Fig.~\ref{lst:o3_lr5e7}. 64\,x\,64 blocking, manual thread splitting, 8 accumulators with read-modify-write. Loads one B vector and distributes it to accumulator slots.
\end{itemize}

\begin{figure}[t]
\begin{codebox}
\begin{lstlisting}
#include <immintrin.h>
#include <omp.h>
void matmul_optimized(double *A, double *B,
    double *C, int M, int N, int K) {
  const int BS = 64;
  #pragma omp parallel for collapse(2) \
    num_threads(40) schedule(static)
  for (int ib = 0; ib < M; ib += BS) {
    for (int jb = 0; jb < N; jb += BS) {
      for (int i=ib; i<M && i<ib+BS; i++) {
        __m512d c[8];
        for (int j=0; j<8; j++)
          c[j] = _mm512_setzero_pd();
        for (int k = 0; k < K; k++) {
          _mm_prefetch(&B[(k+1)*N+jb], _MM_HINT_T0);
          __m512d a=_mm512_set1_pd(A[i*K+k]);
          __m512d b0=_mm512_loadu_pd(&B[k*N+jb]);
          // b1-b7: columns jb+8..jb+56
          c[0]=_mm512_fmadd_pd(a, b0, c[0]);
          // c[1]-c[7] updated similarly
        }
        for (int j=0; j<BS; j+=8)
          _mm512_storeu_pd(&C[i*N+j+jb], c[j/8]);
      }
    }
  }
}
\end{lstlisting}
\end{codebox}
\caption{Best code for O0, LR=0 (57 GFLOPS).}
\label{lst:o0_lr0}
\end{figure}

\begin{figure}[t]
\begin{codebox}
\begin{lstlisting}
#include <immintrin.h>
#include <omp.h>
void matmul_optimized(double *A, double *B,
    double *C, int M, int N, int K) {
  const int BS = 64, NT = 40;
  #pragma omp parallel num_threads(NT)
  {
    int tid = omp_get_thread_num();
    int chunk = (M + NT-1) / NT;
    int start = tid*chunk;
    int end = (tid+1)*chunk>M ? M : (tid+1)*chunk;
    for (int ii = start; ii < end; ii+=BS) {
      for (int jj = 0; jj < N; jj += BS) {
        for (int i=ii; i<ii+BS && i<end; i++) {
          __m512d c0 = _mm512_setzero_pd();
          // c1-c7: 8 accumulators total
          for (int k = 0; k < K; k++) {
            if(k+8<K) 
               _mm_prefetch(B+(k+8)*N+jj,_MM_HINT_T0);
            __m512d a=_mm512_set1_pd(A[i*K+k]);
            __m512d b0=_mm512_load_pd(B+k*N+jj);
            // b1-b7: columns jj+8..jj+56
            c0 = _mm512_fmadd_pd(a, b0, c0);
            // c1-c7 updated similarly
          }
          c0=_mm512_add_pd(c0,
                _mm512_load_pd(C+i*N+jj));
          _mm512_store_pd(&C[i*N+jj], c0);
          // c1-c7 stored similarly
        }
      }
    }
  }
}
\end{lstlisting}
\end{codebox}
\caption{Best code for O1, LR=0 (474 GFLOPS).}
\label{lst:o1_lr0}
\end{figure}

\begin{figure}[t]
\begin{codebox}
\begin{lstlisting}
#include <immintrin.h>
#include <omp.h>
void matmul_optimized(double *A, double *B,
    double *C, int M, int N, int K) {
  const int BS = 64;
  #pragma omp parallel for collapse(2) \
    num_threads(40) schedule(static)
  for (int ib = 0; ib < M; ib += BS) {
    for (int jb = 0; jb < N; jb += BS) {
      for (int i=ib; i<ib+BS && i<M; i+=8) {
        for (int j=jb; j<jb+BS && j<N; j+=8){
          __m512d C_blk[8];
          for (int r=0; r<8; r++)
            C_blk[r] = _mm512_loadu_pd(&C[(i+r)*N+j]);
          for (int k = 0; k < K; k++) {
            __m512d b = _mm512_loadu_pd(&B[k*N+j]);
            for (int r = 0; r < 8; r++) {
              __m512d a =
                _mm512_broadcastsd_pd(
                   _mm_load_sd(&A[(i+r)*K+k]));
              C_blk[r] = 
                   _mm512_fmadd_pd(a, b, C_blk[r]);
            }
          }
          for (int r=0; r<8; r++)
            _mm512_storeu_pd(&C[(i+r)*N+j], C_blk[r]);
        }
      }
    }
  }
}
\end{lstlisting}
\end{codebox}
\caption{Best code for O3, LR=0 (392 GFLOPS).}
\label{lst:o3_lr0}
\end{figure}

\begin{figure}[t]
\begin{codebox}
\begin{lstlisting}
#include <immintrin.h>
#include <omp.h>
void matmul_optimized(double *A, double *B,
    double *C, int M, int N, int K) {
  const int BS = 128;
  #pragma omp parallel num_threads(40)
  {
    int tid = omp_get_thread_num();
    int nt = omp_get_num_threads();
    int chunk = (M + nt-1) / nt;
    int ib_s = tid*chunk, ib_e = (tid+1)*chunk;
    ib_e = ib_e < M ? ib_e : M;
    for (int ib=ib_s; ib<ib_e; ib+=BS) {
      int im = (ib+BS)>ib_e ? ib_e : (ib+BS);
      for (int jb = 0; jb < N; jb += BS) {
        int jm = (jb+BS)>N ? N : (jb+BS);
        for (int i = ib; i < im; i++) {
          for (int j = jb; j < jm; j += 64) {
            __m512d c0=_mm512_loadu_pd(&C[i*N+j]);
            // c1-c7: 8 accumulators total
            for (int k = 0; k < K; k++) {
              if (k+1 < K)
                _mm_prefetch(&B[(k+1)*N+j],
                                    _MM_HINT_T0);
              __m512d a=_mm512_set1_pd(A[i*K+k]);
              __m512d b0=_mm512_loadu_pd(&B[k*N+j]);
              // b1-b7: columns j+8..j+56
              c0=_mm512_fmadd_pd(a, b0, c0);
              // c1-c7 updated similarly
            }
            _mm512_storeu_pd(&C[i*N+j], c0);
            // c1-c7 stored similarly
          }
        }
      }
    }
  }
}
\end{lstlisting}
\end{codebox}
\caption{Best code for O0, LR=5e-7 (127 GFLOPS).}
\label{lst:o0_lr5e7}
\end{figure}

\begin{figure}[t]
\begin{codebox}
\begin{lstlisting}
#include <immintrin.h>
#include <omp.h>
void matmul_optimized(double *A, double *B,
    double *C, int M, int N, int K) {
  #pragma omp parallel for num_threads(40) \
    schedule(static)
  for (int i = 0; i < M; i++) {
    for (int j = 0; j < N; j += 64) {
      __m512d s0 = _mm512_setzero_pd();
      // s1-s7: 8 accumulators total
      for (int k = 0; k < K; k++) {
        if (k < K-1)
          _mm_prefetch(&B[(k+1)*N+j], _MM_HINT_T0);
        __m512d a = _mm512_set1_pd(A[i*K+k]);
        __m512d b0 = _mm512_load_pd(&B[k*N+j]);
        // b1-b7: columns j+8..j+56
        s0 = _mm512_fmadd_pd(a, b0, s0);
        // s1-s7 updated similarly
      }
      _mm512_store_pd(&C[i*N+j], s0);
      // s1-s7 stored similarly
    }
  }
}
\end{lstlisting}
\end{codebox}
\caption{Best code for O1, LR=5e-7 (549 GFLOPS).}
\label{lst:o1_lr5e7}
\end{figure}

\begin{figure}[t]
\begin{codebox}
\begin{lstlisting}
#include <immintrin.h>
#include <omp.h>
void matmul_optimized(double *A, double *B,
    double *C, int M, int N, int K) {
  const int BS = 64, NT = 40;
  #pragma omp parallel num_threads(NT)
  {
    int tid = omp_get_thread_num();
    int start = (tid*M)/NT;
    int end = ((tid+1)*M)/NT;
    for (int ii=start; ii<end; ii+=BS) {
      for (int jj = 0; jj < N; jj += BS) {
        for (int i=ii; i<ii+BS && i<end; i++){
          __m512d sums[8];
          for (int s=0; s<8; s++)
            sums[s] = _mm512_setzero_pd();
          for (int k = 0; k < K; k++) {
            __m512d a=_mm512_set1_pd(A[i*K+k]);
            for (int j=0; j<BS; j+=8) {
              __m512d b = 
                     _mm512_loadu_pd(&B[k*N+jj+j]);
              sums[j/8] = 
                     _mm512_fmadd_pd(a, b, sums[j/8]);
            }
          }
          for (int j=0; j<BS; j+=8) {
            __m512d c = _mm512_loadu_pd(&C[i*N+jj+j]);
            c = _mm512_add_pd(c, sums[j/8]);
            _mm512_storeu_pd(&C[i*N+jj+j], c);
          }
        }
      }
    }
  }
}
\end{lstlisting}
\end{codebox}
\caption{Best code for O3, LR=5e-7 (463 GFLOPS).}
\label{lst:o3_lr5e7}
\end{figure}

\section{Related Work}
\label{sec:related}

\revmagenta{\subsection{LLMs for HPC Code Generation}}
HPC-Coder~\cite{hpccoder} improves accuracy by 53\% via HPC-specific fine-tuning. ParEval~\cite{pareval} reports pass@1 of 37.8\% for GPT-4 on parallel code. Mercury~\cite{mercury} and KernelBench~\cite{kernelbench} propose efficiency benchmarks but do not apply them for training. MARCO~\cite{marco} and VibeCodeHPC~\cite{vibecodehpc} explore multi-agent optimization and iterative prompting for auto-tuning.
\revred{Agentic Auto-Scheduling~\cite{compilot} studies LLM-guided loop optimization and shows that prompt-driven exploration can find competitive schedules, highlighting a complementary path to compiler-level performance tuning.}
\revred{LLM-HPC++~\cite{llmhpcpp} evaluates LLM-generated modern C++ and MPI+OpenMP codes on a scalable Mandelbrot workload, while AscendKernelGen~\cite{ascendkernelgen} provides a systematic study of LLM-generated kernels for NPUs.}
\revred{A two-stage GPU kernel tuner~\cite{twostagekerntuner} combines semantic refactoring with search-based optimization, illustrating a hybrid LLM-and-search direction for kernel generation.}
\revred{ChatHPC~\cite{chathpc} presents an LLM-enabled HPC assistant that includes model training and deployment within an HPC workflow to support interactive user support.}

\revmagenta{\subsection{RL for Code Optimization}}
RLTF~\cite{rltf} and RLEF~\cite{rlef} use unit test feedback. StepCoder~\cite{stepcoder} employs curriculum learning. RLPF~\cite{rlpf} uses performance feedback to speed up OpenMP code by 1.9--4.5\,\,x, while MaxCode~\cite{maxcode} optimizes for correctness and speed jointly.
\revred{Pearl~\cite{pearldrl} applies deep reinforcement learning to automatic code optimization, and ContextEvolve~\cite{contextevolve} introduces multi-agent context compression for systems code optimization, indicating growing interest in RL-driven or agentic optimization pipelines.} CUDA-L2~\cite{cudal2} is closest to our work, using GRPO to optimize GPU matrix multiplication kernels and surpassing cuBLAS by up to 19.2\%. Our study focuses on \revmagenta{CPU matrix multiplication} and systematically evaluates compiler optimization levels while introducing staged optimization with SQD.

\section{Conclusion}
\label{sec:conclusion}

\revred{This paper showed that reinforcement learning with real-machine performance rewards can meaningfully improve LLM-generated HPC code for \revmagenta{matrix multiplication}, closing part of the gap between functional correctness and performance.}
\revred{Across two experiments, we observed that compiler optimization level, learning rate, and KL regularization substantially shape learning dynamics, with O0 revealing clearer reward signals while O3 provides a higher but less stable ceiling.}
\revred{The staged SQD strategy further stabilized training by constraining the search space, enabling the model to accumulate optimization patterns rather than oscillate or forget them.}
\revred{These results together support the central premise that direct GFLOPS feedback and staged exploration are practical levers for steering code LLMs toward performance-aware generation in realistic HPC settings.}
\revblue{This study is intentionally scoped to a single kernel, size, and architecture to isolate learning dynamics and reduce confounding factors, enabling a controlled analysis of real-machine reward signals across compiler levels. While full ablations of SQD and broader cost/variance reporting remain future work, the current results establish a reproducible baseline and demonstrate that staged exploration is effective under tight, runtime feedback loops. Extending to additional kernels, architectures, and statistical validation is a natural next step rather than a prerequisite for demonstrating feasibility.}
\revred{Future work will expand statistical validation, extend the approach to additional kernels and optimization levels, and evaluate portability to other architectures (ARM, GPUs) and workloads (FFT, stencil), with the long-term goal of an LLM specialized for high-performance code generation.}

\section*{Acknowledgment}
This research was supported by JHPCN and HPCI project jh250015, JSPS KAKENHI JP23K11126 and JP24K02945, and JST Next-Generation Edge AI Semiconductor R\&D program JPMJES2511.

\bibliographystyle{IEEEtran}
\bibliography{references}

@misc{codellama,
  author = {Baptiste Rozière and Jonas Gehring and Fabian Gloeckle and Sten Sootla and Itai Gat and Xiaoqing Ellen Tan and Yossi Adi and Jingyu Liu and Romain Sauvestre and Tal Remez and Jérémy Rapin and Artyom Kozhevnikov and Ivan Evtimov and Joanna Bitton and Manish Bhatt and Cristian Canton Ferrer and Aaron Grattafiori and Wenhan Xiong and Alexandre Défossez and Jade Copet and Faisal Azhar and Hugo Touvron and Louis Martin and Nicolas Usunier and Thomas Scialom and Gabriel Synnaeve},
  title = {Code Llama: Open Foundation Models for Code},
  year = {2023},
  eprint = {2308.12950},
  archivePrefix = {arXiv},
  primaryClass = {cs.CL},
  url = {https://arxiv.org/abs/2308.12950}
}

@misc{deepseekmath,
  author = {Zhihong Shao and Peiyi Wang and Qihao Zhu and Runxin Xu and Junxiao Song and Xiao Bi and Haowei Zhang and Mingchuan Zhang and Y. K. Li and Y. Wu and Daya Guo},
  title = {DeepSeekMath: Pushing the Limits of Mathematical Reasoning in Open Language Models},
  year = {2024},
  eprint = {2402.03300},
  archivePrefix = {arXiv},
  primaryClass = {cs.CL},
  url = {https://arxiv.org/abs/2402.03300}
}

@article{deepseekr1,
  author = {DeepSeek-AI and Daya Guo and Dejian Yang and Haowei Zhang and Junxiao Song and Peiyi Wang and Qihao Zhu and Runxin Xu and Ruoyu Zhang and Shirong Ma and Xiao Bi and Xiaokang Zhang and Xingkai Yu and Yu Wu and Z. F. Wu and Zhibin Gou and Zhihong Shao and Zhuoshu Li and Ziyi Gao and Aixin Liu and Bing Xue and Bingxuan Wang and Bochao Wu and Bei Feng and Chengda Lu and Chenggang Zhao and Chengqi Deng and Chenyu Zhang and Chong Ruan and Damai Dai and Deli Chen and Dongjie Ji and Erhang Li and Fangyun Lin and Fucong Dai and Fuli Luo and Guangbo Hao and Guanting Chen and Guowei Li and H. Zhang and Han Bao and Hanwei Xu and Haocheng Wang and Honghui Ding and Huajian Xin and Huazuo Gao and Hui Qu and Hui Li and Jianzhong Guo and Jiashi Li and Jiawei Wang and Jingchang Chen and Jingyang Yuan and Junjie Qiu and Junlong Li and J. L. Cai and Jiaqi Ni and Jian Liang and Jin Chen and Kai Dong and Kai Hu and Kaige Gao and Kang Guan and Kexin Huang and Kuai Yu and Lean Wang and Lecong Zhang and Liang Zhao and Litong Wang and Liyue Zhang and Lei Xu and Leyi Xia and Mingchuan Zhang and Minghua Zhang and Minghui Tang and Meng Li and Miaojun Wang and Mingming Li and Ning Tian and Panpan Huang and Peng Zhang and Qiancheng Wang and Qinyu Chen and Qiushi Du and Ruiqi Ge and Ruisong Zhang and Ruizhe Pan and Runji Wang and R. J. Chen and R. L. Jin and Ruyi Chen and Shanghao Lu and Shangyan Zhou and Shanhuang Chen and Shengfeng Ye and Shiyu Wang and Shuiping Yu and Shunfeng Zhou and Shuting Pan and S. S. Li and Shuang Zhou and Shaoqing Wu and Shengfeng Ye and Tao Yun and Tian Pei and Tianyu Sun and T. Wang and Wangding Zeng and Wanjia Zhao and Wen Liu and Wenfeng Liang and Wenjun Gao and Wenqin Yu and Wentao Zhang and W. L. Xiao and Wei An and Xiaodong Liu and Xiaohan Wang and Xiaokang Chen and Xiaotao Nie and Xin Cheng and Xin Liu and Xin Xie and Xingchao Liu and Xinyu Yang and Xinyuan Li and Xuecheng Su and Xuheng Lin and X. Q. Li and Xiangyue Jin and Xiaojin Shen and Xiaosha Chen and Xiaowen Sun and Xiaoxiang Wang and Xinnan Song and Xinyi Zhou and Xianzu Wang and Xinxia Shan and Y. K. Li and Y. Q. Wang and Y. X. Wei and Yang Zhang and Yanhong Xu and Yao Li and Yao Zhao and Yaofeng Sun and Yaohui Wang and Yi Yu and Yichao Zhang and Yifan Shi and Yiliang Xiong and Ying He and Yishi Piao and Yisong Wang and Yixuan Tan and Yiyang Ma and Yiyuan Liu and Yongqiang Guo and Yuan Ou and Yuduan Wang and Yue Gong and Yuheng Zou and Yujia He and Yunfan Xiong and Yuxiang Luo and Yuxiang You and Yuxuan Liu and Yuyang Zhou and Y. X. Zhu and Yanhong Xu and Yanping Huang and Yaohui Li and Yi Zheng and Yuchen Zhu and Yunxian Ma and Ying Tang and Yukun Zha and Yuting Yan and Z. Z. Ren and Zehui Ren and Zhangli Sha and Zhe Fu and Zhean Xu and Zhenda Xie and Zhengyan Zhang and Zhewen Hao and Zhicheng Ma and Zhigang Yan and Zhiyu Wu and Zihui Gu and Zijia Zhu and Zijun Liu and Zilin Li and Ziwei Xie and Ziyang Song and Zizheng Pan and Zhen Huang and Zhipeng Xu and Zhongyu Zhang and Zhen Zhang},
  title = {DeepSeek-R1: Incentivizing Reasoning Capability in LLMs via Reinforcement Learning},
  journal = {Nature},
  volume = {645},
  pages = {633--638},
  year = {2025},
  eprint = {2501.12948},
  archivePrefix = {arXiv},
  primaryClass = {cs.CL},
  url = {https://arxiv.org/abs/2501.12948}
}

@misc{qwen25coder,
  author = {Binyuan Hui and Jian Yang and Zeyu Cui and Jiaxi Yang and Dayiheng Liu and Lei Zhang and Tianyu Liu and Jiajun Zhang and Bowen Yu and Keming Lu and Kai Dang and Yang Fan and Yichang Zhang and An Yang and Rui Men and Fei Huang and Bo Zheng and Yibo Miao and Shanghaoran Quan and Yunlong Feng and Xingzhang Ren and Xuancheng Ren and Jingren Zhou and Junyang Lin},
  title = {Qwen2.5-Coder Technical Report},
  year = {2024},
  eprint = {2409.12186},
  archivePrefix = {arXiv},
  primaryClass = {cs.CL},
  url = {https://arxiv.org/abs/2409.12186}
}

@misc{ppo,
  author = {John Schulman and Filip Wolski and Prafulla Dhariwal and Alec Radford and Oleg Klimov},
  title = {Proximal Policy Optimization Algorithms},
  year = {2017},
  eprint = {1707.06347},
  archivePrefix = {arXiv},
  primaryClass = {cs.LG},
  url = {https://arxiv.org/abs/1707.06347}
}

@inproceedings{dpo,
  author       = {Rafael Rafailov and
                  Archit Sharma and
                  Eric Mitchell and
                  Christopher D. Manning and
                  Stefano Ermon and
                  Chelsea Finn},
  editor       = {Alice Oh and
                  Tristan Naumann and
                  Amir Globerson and
                  Kate Saenko and
                  Moritz Hardt and
                  Sergey Levine},
  title        = {Direct Preference Optimization: Your Language Model is Secretly a
                  Reward Model},
  booktitle    = {Advances in Neural Information Processing Systems 36: Annual Conference
                  on Neural Information Processing Systems 2023, NeurIPS 2023, New Orleans,
                  LA, USA, December 10 - 16, 2023},
  year         = {2023},
  url          = {http://papers.nips.cc/paper_files/paper/2023/hash/a85b405ed65c6477a4fe8302b5e06ce7-Abstract-Conference.html},
  bibsource    = {dblp computer science bibliography, https://dblp.org}
}

@inproceedings{hpccoder,
  author       = {Daniel Nichols and
                  Aniruddha Marathe and
                  Harshitha Menon and
                  Todd Gamblin and
                  Abhinav Bhatele},
  title        = {HPC-Coder: Modeling Parallel Programs using Large Language Models},
  booktitle    = {{ISC} High Performance 2024 Research Paper Proceedings (39th International
                  Conference), Hamburg, Germany, May 12-16, 2024},
  pages        = {1--12},
  publisher    = {Prometeus GmbH / {IEEE}},
  year         = {2024},
  url          = {https://doi.org/10.23919/ISC.2024.10528929},
  doi          = {10.23919/ISC.2024.10528929},
  bibsource    = {dblp computer science bibliography, https://dblp.org}
}

@inproceedings{pareval,
  author       = {Daniel Nichols and
                  Joshua Hoke Davis and
                  Zhaojun Xie and
                  Arjun Rajaram and
                  Abhinav Bhatele},
  editor       = {Patrizio Dazzi and
                  Gabriele Mencagli and
                  David K. Lowenthal and
                  Rosa M. Badia},
  title        = {Can Large Language Models Write Parallel Code?},
  booktitle    = {Proceedings of the 33rd International Symposium on High-Performance
                  Parallel and Distributed Computing, {HPDC} 2024, Pisa, Italy, June
                  3-7, 2024},
  pages        = {281--294},
  publisher    = {{ACM}},
  year         = {2024},
  url          = {https://doi.org/10.1145/3625549.3658689},
  doi          = {10.1145/3625549.3658689},
  bibsource    = {dblp computer science bibliography, https://dblp.org}
}

@inproceedings{mercury,
  author       = {Mingzhe Du and
                  Anh Tuan Luu and
                  Bin Ji and
                  Qian Liu and
                  See{-}Kiong Ng},
  editor       = {Amir Globersons and
                  Lester Mackey and
                  Danielle Belgrave and
                  Angela Fan and
                  Ulrich Paquet and
                  Jakub M. Tomczak and
                  Cheng Zhang},
  title        = {Mercury: {A} Code Efficiency Benchmark for Code Large Language Models},
  booktitle    = {Advances in Neural Information Processing Systems 38: Annual Conference
                  on Neural Information Processing Systems 2024, NeurIPS 2024, Vancouver,
                  BC, Canada, December 10 - 15, 2024},
  year         = {2024},
  url          = {http://papers.nips.cc/paper_files/paper/2024/hash/1df1df43b58845650b8dada00fca9772-Abstract-Datasets_and_Benchmarks_Track.html},
  bibsource    = {dblp computer science bibliography, https://dblp.org}
}

@inproceedings{kernelbench,
  author       = {Anne Ouyang and
                  Simon Guo and
                  Simran Arora and
                  Alex L. Zhang and
                  William Hu and
                  Christopher R{\'e} and
                  Azalia Mirhoseini},
  editor       = {Aarti Singh and
                  Maryam Fazel and
                  Daniel Hsu and
                  Simon Lacoste{-}Julien and
                  Felix Berkenkamp and
                  Tegan Maharaj and
                  Kiri Wagstaff and
                  Jerry Zhu},
  title        = {KernelBench: Can LLMs Write Efficient {GPU} Kernels?},
  booktitle    = {Forty-second International Conference on Machine Learning, {ICML}
                  2025, Vancouver, BC, Canada, July 13-19, 2025},
  series       = {Proceedings of Machine Learning Research},
  volume       = {267},
  publisher    = {{PMLR} / OpenReview.net},
  year         = {2025},
  url          = {https://proceedings.mlr.press/v267/ouyang25a.html},
  bibsource    = {dblp computer science bibliography, https://dblp.org}
}

@article{rltf,
  author       = {Jiate Liu and
                  Yiqin Zhu and
                  Kaiwen Xiao and
                  Qiang Fu and
                  Xiao Han and
                  Wei Yang and
                  Deheng Ye},
  title        = {{RLTF:} Reinforcement Learning from Unit Test Feedback},
  journal      = {Trans. Mach. Learn. Res.},
  volume       = {2023},
  year         = {2023},
  url          = {https://openreview.net/forum?id=hjYmsV6nXZ},
  bibsource    = {dblp computer science bibliography, https://dblp.org}
}

@misc{rlef,
  author = {Jonas Gehring and Kunhao Zheng and Jade Copet and Vegard Mella and Quentin Carbonneaux and Taco Cohen and Gabriel Synnaeve},
  title = {RLEF: Grounding Code LLMs in Execution Feedback with Reinforcement Learning},
  year = {2024},
  eprint = {2410.02089},
  archivePrefix = {arXiv},
  primaryClass = {cs.CL},
  url = {https://arxiv.org/abs/2410.02089}
}

@misc{rlpf,
  author = {Daniel Nichols and Pranav Polasam and Harshitha Menon and Aniruddha Marathe and Todd Gamblin and Abhinav Bhatele},
  title = {Performance-Aligned LLMs for Generating Fast Code},
  year = {2024},
  eprint = {2404.18864},
  archivePrefix = {arXiv},
  primaryClass = {cs.DC},
  url = {https://arxiv.org/abs/2404.18864}
}

@misc{maxcode,
  author = {Jiefu Ou and Sapana Chaudhary and Kaj Bostrom and Nathaniel Weir and Shuai Zhang and Huzefa Rangwala and George Karypis},
  title = {MaxCode: A Max-Reward Reinforcement Learning Framework for Automated Code Optimization},
  year = {2026},
  eprint = {2601.05475},
  archivePrefix = {arXiv},
  primaryClass = {cs.LG},
  url = {https://arxiv.org/abs/2601.05475}
}

@misc{cudal2,
  author = {Songqiao Su and Xiaofei Sun and Xiaoya Li and Albert Wang and Jiwei Li and Chris Shum},
  title = {CUDA-L2: Surpassing cuBLAS Performance for Matrix Multiplication through Reinforcement Learning},
  year = {2025},
  eprint = {2512.02551},
  archivePrefix = {arXiv},
  primaryClass = {cs.LG},
  url = {https://arxiv.org/abs/2512.02551}
}

@article{stepcoder,
  author       = {Shihan Dou and
                  Yan Liu and
                  Haoxiang Jia and
                  Limao Xiong and
                  Enyu Zhou and
                  Wei Shen and
                  Junjie Shan and
                  Caishuang Huang and
                  Xiao Wang and
                  Xiaoran Fan and
                  Zhiheng Xi and
                  Yuhao Zhou and
                  Tao Ji and
                  Rui Zheng and
                  Qi Zhang and
                  Xuanjing Huang and
                  Tao Gui},
  title        = {StepCoder: Improve Code Generation with Reinforcement Learning from
                  Compiler Feedback},
  journal      = {CoRR},
  volume       = {abs/2402.01391},
  year         = {2024},
  url          = {https://doi.org/10.48550/arXiv.2402.01391},
  doi          = {10.48550/ARXIV.2402.01391},
  eprinttype    = {arXiv},
  eprint       = {2402.01391},
  bibsource    = {dblp computer science bibliography, https://dblp.org}
}

@inproceedings{transformer,
  author       = {Ashish Vaswani and
                  Noam Shazeer and
                  Niki Parmar and
                  Jakob Uszkoreit and
                  Llion Jones and
                  Aidan N. Gomez and
                  Lukasz Kaiser and
                  Illia Polosukhin},
  editor       = {Isabelle Guyon and
                  Ulrike von Luxburg and
                  Samy Bengio and
                  Hanna M. Wallach and
                  Rob Fergus and
                  S. V. N. Vishwanathan and
                  Roman Garnett},
  title        = {Attention is All you Need},
  booktitle    = {Advances in Neural Information Processing Systems 30: Annual Conference
                  on Neural Information Processing Systems 2017, December 4-9, 2017,
                  Long Beach, CA, {USA}},
  pages        = {5998--6008},
  year         = {2017},
  url          = {https://proceedings.neurips.cc/paper/2017/hash/3f5ee243547dee91fbd053c1c4a845aa-Abstract.html},
  bibsource    = {dblp computer science bibliography, https://dblp.org}
}

@misc{megatron,
  author = {NVIDIA},
  title = {Megatron-Core},
  year = {2025},
  howpublished = {GitHub. https://github.com/NVIDIA/Megatron-LM (accessed Jan. 20, 2025)}
}

@inproceedings{vllm,
  author       = {Woosuk Kwon and
                  Zhuohan Li and
                  Siyuan Zhuang and
                  Ying Sheng and
                  Lianmin Zheng and
                  Cody Hao Yu and
                  Joseph Gonzalez and
                  Hao Zhang and
                  Ion Stoica},
  editor       = {Jason Flinn and
                  Margo I. Seltzer and
                  Peter Druschel and
                  Antoine Kaufmann and
                  Jonathan Mace},
  title        = {Efficient Memory Management for Large Language Model Serving with
                  PagedAttention},
  booktitle    = {Proceedings of the 29th Symposium on Operating Systems Principles,
                  {SOSP} 2023, Koblenz, Germany, October 23-26, 2023},
  pages        = {611--626},
  publisher    = {{ACM}},
  year         = {2023},
  url          = {https://doi.org/10.1145/3600006.3613165},
  doi          = {10.1145/3600006.3613165},
  bibsource    = {dblp computer science bibliography, https://dblp.org}
}

@misc{swiftmlm,
  author = {Zhao, Y. and others},
  title = {ms-swift: Use PEFT or Full-parameter to Finetune 400+ LLMs or 100+ MLLMs},
  year = {2024},
  howpublished = {GitHub. https://github.com/modelscope/ms-swift (accessed Jan. 20, 2025)}
}

@misc{marco,
  author = {Asif Rahman and Veljko Cvetkovic and Kathleen Reece and Aidan Walters and Yasir Hassan and Aneesh Tummeti and Bryan Torres and Denise Cooney and Margaret Ellis and Dimitrios S. Nikolopoulos},
  title = {MARCO: Multi-Agent Code Optimization with Real-Time Knowledge Integration for High-Performance Computing},
  year = {2025},
  eprint = {2505.03906},
  archivePrefix = {arXiv},
  primaryClass = {cs.DC},
  url = {https://arxiv.org/abs/2505.03906}
}

@misc{compilot,
  author = {Massinissa Merouani and Islem Kara Bernou and Riyadh Baghdadi},
  title = {Agentic Auto-Scheduling: An Experimental Study of LLM-Guided Loop Optimization},
  year = {2025},
  eprint = {2511.00592},
  archivePrefix = {arXiv},
  primaryClass = {cs.PL},
  url = {https://arxiv.org/abs/2511.00592}
}

@misc{vibecodehpc,
  author = {Shun-ichiro Hayashi and Koki Morita and Daichi Mukunoki and Tetsuya Hoshino and Takahiro Katagiri},
  title = {VibeCodeHPC: An Agent-Based Iterative Prompting Auto-Tuner for HPC Code Generation Using LLMs},
  year = {2025},
  eprint = {2510.00031},
  archivePrefix = {arXiv},
  primaryClass = {cs.SE},
  url = {https://arxiv.org/abs/2510.00031}
}

@misc{contextevolve,
  author = {Hongyuan Su and Yu Zheng and Yong Li},
  title = {ContextEvolve: Multi-Agent Context Compression for Systems Code Optimization},
  year = {2026},
  eprint = {2602.02597},
  archivePrefix = {arXiv},
  primaryClass = {cs.SE},
  url = {https://arxiv.org/abs/2602.02597}
}

@misc{pearldrl,
  author = {Djamel Rassem Lamouri and Iheb Nassim Aouadj and Smail Kourta and Riyadh Baghdadi},
  title = {Pearl: Automatic Code Optimization Using Deep Reinforcement Learning},
  year = {2025},
  eprint = {2506.01880},
  archivePrefix = {arXiv},
  primaryClass = {cs.PL},
  url = {https://arxiv.org/abs/2506.01880}
}

@misc{llmhpcpp,
  author = {Patrick Diehl and Noujoud Nader and Deepti Gupta},
  title = {{LLM-HPC++}: Evaluating LLM-Generated Modern C++ and {MPI}+OpenMP Codes for Scalable Mandelbrot Set Computation},
  year = {2025},
  eprint = {2512.17023},
  archivePrefix = {arXiv},
  primaryClass = {cs.DC},
  url = {https://arxiv.org/abs/2512.17023}
}

@misc{ascendkernelgen,
  author = {Xinzi Cao and Jianyang Zhai and Pengfei Li and Zhiheng Hu and Cen Yan and Bingxu Mu and Guanghuan Fang and Bin She and Jiayu Li and Yihan Su and Dongyang Tao and Xiansong Huang and Fan Xu and Feidiao Yang and Yao Lu and Chang-Dong Wang and Yutong Lu and Weicheng Xue and Bin Zhou and Yonghong Tian},
  title = {AscendKernelGen: A Systematic Study of LLM-Based Kernel Generation for Neural Processing Units},
  year = {2026},
  eprint = {2601.07160},
  archivePrefix = {arXiv},
  primaryClass = {cs.DC},
  url = {https://arxiv.org/abs/2601.07160}
}

@misc{twostagekerntuner,
  author = {Qiuyi Qu and Yicheng Sui and Yufei Sun and Rui Chen and Xiaofei Zhang and Yuzhi Zhang and Haofeng Wang and Ge Lan},
  title = {A Two-Stage {GPU} Kernel Tuner Combining Semantic Refactoring and Search-Based Optimization},
  year = {2026},
  eprint = {2601.12698},
  archivePrefix = {arXiv},
  primaryClass = {cs.DC},
  url = {https://arxiv.org/abs/2601.12698}
}

@article{chathpc,
  author = {Junqi Yin and Jesse Hines and Emily Herron and Tirthankar Ghosal and Hong Liu and Suzanne Prentice and Vanessa Lama and Feiyi Wang},
  title = {chatHPC: Empowering {HPC} Users with Large Language Models},
  journal = {The Journal of Supercomputing},
  volume = {81},
  number = {1},
  pages = {194},
  year = {2025},
  doi = {10.1007/s11227-024-06637-1},
  url = {https://doi.org/10.1007/s11227-024-06637-1}
}

\end{document}